\documentclass{article}

\usepackage{PRIMEarxiv}

\usepackage[utf8]{inputenc} % allow utf-8 input
\usepackage[T1]{fontenc}    % use 8-bit T1 fonts
\usepackage{hyperref}       % hyperlinks
\usepackage{url}            % simple URL typesetting
\usepackage{booktabs}       % professional-quality tables
\usepackage{amsfonts}       % blackboard math symbols
\usepackage{nicefrac}       % compact symbols for 1/2, etc.
\usepackage{microtype}      % microtypography
\usepackage{fancyhdr}       % header
\usepackage{graphicx}       % graphics
\graphicspath{{figures/}{examples/}{media/}}     % organize your images and other figures under media/ folder
\usepackage{amsmath}
\usepackage{adjustbox}
\usepackage[table]{xcolor}
\usepackage{wrapfig}
\usepackage{multirow}
\newcommand{\bestA}[1]{\cellcolor{blue!12}\textbf{#1}}  % ≥ group best
\title{LightOnOCR: A 1B End-to-End Multilingual Vision-Language Model for State-of-the-Art OCR
}

\author{
  Said Taghadouini \quad Adrien Cavaill\`es \quad Baptiste Aubertin \\
  LightOn
}

\begin{document}
\maketitle
\thispagestyle{empty}

\begin{abstract}
We present \textbf{LightOnOCR-2-1B}, a 1B-parameter end-to-end multilingual vision--language model that converts document images (e.g., PDFs) into clean, naturally ordered text without brittle OCR pipelines. Trained on a large-scale, high-quality distillation mix with strong coverage of scans, French documents, and scientific PDFs, LightOnOCR-2 achieves state-of-the-art results on OlmOCR-Bench while being 9$\times$ smaller and substantially faster than prior best-performing models. We further extend the output format to predict normalized bounding boxes for embedded images, introducing localization during pretraining via a resume strategy and refining it with RLVR using IoU-based rewards. Finally, we improve robustness with checkpoint averaging and task-arithmetic merging. We release model checkpoints under Apache 2.0, and publicly release the dataset and \textbf{LightOnOCR-bbox-bench} evaluation under their respective licenses.

\end{abstract}

\begin{center}
\setlength{\tabcolsep}{8pt}
\renewcommand{\arraystretch}{1.1}
\begin{tabular}{@{}rl@{}}
\textbf{Model:}  & \url{https://huggingface.co/collections/lightonai/lightonocr-2} \\
\textbf{Blog:}     & \url{https://huggingface.co/blog/lightonai/lightonocr-2} \\
% \textbf{Datasets:}  & \url{https://huggingface.co/datasets/lightonai/LightOnOCR-mix-0126} \\
%                     & \url{https://huggingface.co/datasets/lightonai/LightOnOCR-bbox-mix-0126} \\
\textbf{Benchmark:}     & \url{https://huggingface.co/datasets/lightonai/LightOnOCR-bbox-bench} \\
\end{tabular}
\end{center}

% keywords can be removed
% \keywords{OCR \and end-to-end \and More}

\section{Introduction}

Despite decades of progress in Optical Character Recognition (OCR), from classical engines such as Tesseract~\cite{smith2007tesseract}, to deep neural sequence recognizers like CRNN~\cite{shi2015crnn}, and Transformer-based models such as TrOCR~\cite{li2021trocr}, real-world documents remain challenging: reading order can be ambiguous in multi-column layouts, tables require consistent structure, and scientific PDFs often mix dense typography with mathematical notation, figures, and noisy scans. A common production strategy is to rely on multi-stage pipelines (e.g., layout analysis, text detection, text recognition, table extraction, and reading-order reconstruction), as in systems such as PaddleOCR~\cite{paddleocr_repo,du2020ppocr} and MinerU~\cite{mineru2024}. While effective in many settings, these pipelines couple multiple components and intermediate representations, making them costly to adapt: improving performance on a new document distribution often requires additional annotations for intermediate tasks (layout regions, table structure, reading order) and coordinated changes across stages.

End-to-end vision-language models (VLMs) reduce this engineering burden by learning extraction directly from pixels to structured text~\cite{Nougat,poznanski2025olmocr,poznanski2025olmocr2,li2025dotsocr,cui2025paddleocrvl,chandra_repo}. This enables continuous improvement and specialization via straightforward fine-tuning, without retooling each stage of a pipeline. In this work, we present \textbf{LightOnOCR-2}, a compact 1B-parameter end-to-end multilingual VLM that achieves state-of-the-art OCR, outperforming substantially larger systems, and extends OCR with document image localization through predicted bounding boxes.

In this context, we release LightOnOCR, a compact, end-to-end model that delivers state-of-the-art document understanding with lightning speed and low cost. Competing systems(including newest releases) often rely on multiple moving parts to boost performance, but this added complexity makes them brittle, difficult to train, and prone to break when adapting to new data or domains. LightOnOCR, on the other hand, is a single unified model — fully differentiable and easy to optimize end-to-end — capable of handling complex layouts such as tables, forms, receipts, and scientific notation without fragile multi-stage pipelines.

LightOnOCR-2-1B is a compact, end-to-end model that builds on our first release~\cite{blogpostv1} with a substantially expanded and cleaner training mixture ($2.5\times$ larger), with increased coverage of scans, French documents, and scientific content, supported by an improved data curation pipeline. During pretraining, we train at higher resolution (maximum longest edge $1540$px), apply data augmentation, and explicitly include empty pages to reduce looping behaviors and improve full-page fidelity. We then apply Reinforcement Learning with Verifiable Rewards~(RLVR)~\cite{Lambert2024TLU3P} to target persistent failure modes that are difficult to address with supervised learning alone, including repetition loops, math rendering and formatting errors, and layout-sensitive consistency constraints enforced through unit-test style checks. We also train a variant to predict bounding boxes for embedded images. To avoid degrading OCR quality with naive supervised fine-tuning, we introduce coordinate supervision during pretraining via a resume strategy and then refine localization using RLVR with IoU-based objectives. Finally, we leverage lightweight weight-space techniques, checkpoint averaging and task-arithmetic merging, to combine complementary gains and to control the trade-off between OCR quality and bounding box~(bbox) accuracy.

\paragraph{Contributions}
To summarize, we make the following contributions:
\begin{itemize}
    \item We release \textbf{LightOnOCR-2-1B}, a compact 1B-parameter end-to-end multilingual VLM for document parsing that achieves a new state-of-the-art result on OlmOCR-Bench, outperforming substantially larger models (e.g., 9B-scale baselines).
    \item We scale and improve the pretraining mixture ($2.5\times$ larger) with stronger coverage of French and scientific documents, higher-resolution training (max longest edge $1540$px), improved normalization (cleaner \LaTeX{} handling), and robustness via data augmentations and explicit empty-page targets.
    \item We introduce a \texttt{nvpdftex}-based arXiv curation pipeline to obtain pixel-aligned supervision from \TeX{} sources, and use it both to strengthen scientific OCR supervision and to generate an automatic subset for our localization benchmark.
    \item We add image bounding box prediction in dedicated variants by introducing coordinate supervision during pretraining and refining localization with RLVR using IoU-based objectives.
    \item We introduce \textbf{LightOnOCR-bbox-bench}, a new benchmark for image localization in documents.
    \item We show that lightweight weight-space techniques, checkpoint averaging and task-arithmetic merging, can improve OCR and enable controlled trade-offs between OCR quality and bbox accuracy across released checkpoints.
\end{itemize}

\section{Overview}
\begin{figure}[h]
    \centering
    \includegraphics[width=0.5\linewidth]{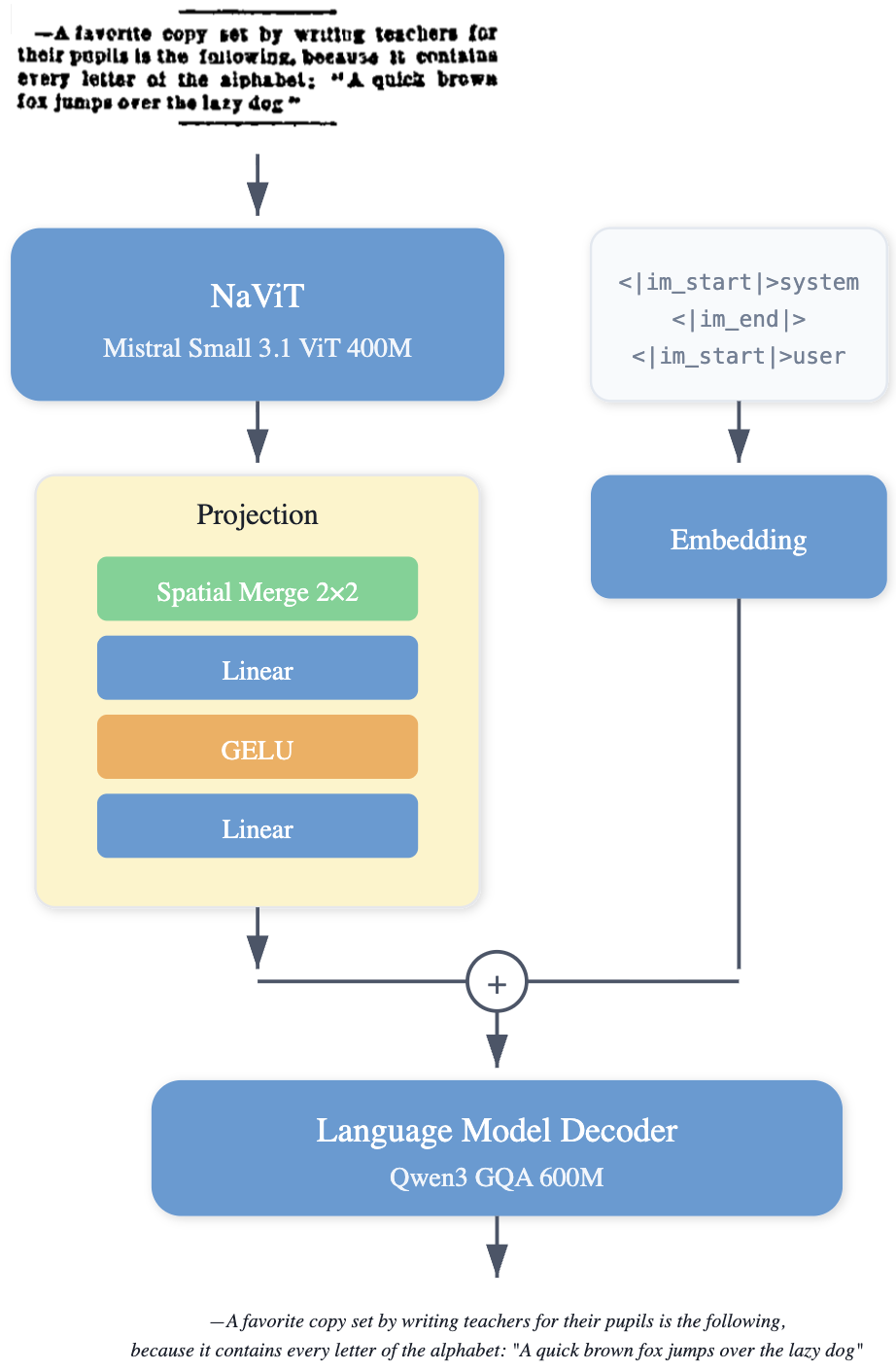}
    \caption{Model architecture. 
    }
    \label{fig:arch}
\end{figure}

In this section, we describe the LightOnOCR architecture and summarize the training recipe differences between LightOnOCR-1B and LightOnOCR-2-1B.

\subsection{Architecture}
LightOnOCR is a compact 1B-parameter vision-language model composed of three main components: a vision encoder, a multimodal projector, and a language model decoder. We train the model to perform OCR without task prompts at inference time, so the extraction behavior is embedded in the weights rather than controlled by an explicit prompt.

\paragraph{Vision Encoder}
We employ a native-resolution Vision Transformer initialized from the pretrained \texttt{Mistral-Small-3.1}~\cite{mistral31_blogpost} vision encoder weights. This encoder handles variable image sizes while preserving spatial structure, which is critical for documents with diverse aspect ratios and fine typographic details.

\paragraph{Multimodal Projector}
To bridge the vision and language modalities while controlling sequence length, we use a two-layer MLP with GELU activation that projects vision features into the language model's embedding space. Before projection, we apply spatial merging with a factor of 2, effectively grouping $2 \times 2$ patches and reducing the number of visual tokens by $4\times$. This keeps the overall token count tractable for high-resolution inputs while preserving sufficient spatial granularity. The projector is randomly initialized and trained from scratch.

\paragraph{Language Model Decoder}
We initialize the decoder from pretrained \texttt{Qwen3}~\cite{qwen3technicalreport}. The decoder produces a single, linearized representation of the page that preserves reading order while emitting structured tokens for non-text elements (e.g., \texttt{![image](image\_N.png)}). To simplify the interface between modalities, we remove the image-break and image-end tokens and condition the decoder on a single contiguous block of visual tokens (after spatial merging), followed by text tokens. This yields a compact end-to-end VLM with a consistent generation format across datasets.

\paragraph{Initialization}
By initializing from strong pretrained components, LightOnOCR inherits robust visual representations and multilingual language modeling capabilities, enabling effective transfer to OCR with reduced training cost.

\subsection{LightOnOCR-1}
The first version of LightOnOCR~\cite{blogpostv1}, established the core architecture described above. It was developed using supervised training on the PDF Association~(PDFA) dataset~\cite{pixparse_pdfa_eng_wds}, using transcriptions generated by \texttt{Qwen2-VL-72B-Instruct}~\cite{Qwen2VL} as teacher supervision. Training used a maximum longest-edge resolution of 1024 pixels.

\subsection{LightOnOCR-2}
\label{sec:overview_v2}
LightOnOCR-2 keeps the same architecture but significantly updates both the data and the training recipe. We scale the pretraining mixture from 17M to 43M pages (Section~\ref{sec:data}), with stronger coverage of scanned documents, scientific PDFs, and European languages with an emphasis on French. Supervision quality is improved by leveraging a stronger teacher, \texttt{Qwen3-VL-235B-A22B-Instruct}~\cite{qwen3technicalreport}, and by improving conversion and normalization so that targets better preserve layout and clean \LaTeX{} formatting. We also increase the maximum longest-edge resolution from 1024 to 1540 pixels, improving legibility for small text and dense mathematical notation while retaining dynamic resizing for diverse page sizes. Beyond transcription, we train an image-localization variant that predicts bounding boxes for embedded images by extending the output format with normalized coordinates; to avoid OCR regressions, we introduce coordinate supervision during pretraining via a resume strategy and then refine localization with RLVR using IoU-based objectives (Section~\ref{sec:rl}). Finally, we apply lightweight weight-space techniques, checkpoint averaging and task-arithmetic merging, to combine complementary gains across runs and to control trade-offs between OCR quality and bbox accuracy.

% \paragraph{FP8 Quantization}
% To reduce memory footprint and accelerate inference, we release an FP8-quantized variant of LightOnOCR-2. Following the approach of OlmOCR~\cite{poznanski2025olmocr2}, we quantize only the language model decoder to FP8 while keeping the vision encoder and multimodal projector in their original precision. This selective strategy is motivated by the asymmetric computational profile of VLMs: while the decoder is invoked autoregressively for every generated token, the vision encoder processes each input image only once per forward pass, making decoder quantization the dominant source of efficiency gains with minimal quality impact.

\section{Datasets and Preprocessing}
\label{sec:data}

\subsection{Dataset Scale and Composition}\label{annotations}
Both LightOnOCR versions are trained on large-scale document OCR corpora built primarily through distillation: a strong vision--language teacher produces naturally ordered transcriptions (Markdown with \LaTeX{} spans) from rendered PDF pages. Compared to LightOnOCR-1B, LightOnOCR-2-1B scales up the data mixture and improves supervision quality by upgrading the teacher model from \texttt{Qwen2-VL-72B-Instruct}~\cite{Qwen2VL} to \texttt{Qwen3-VL-235B-A22B-Instruct}, yielding more faithful mathematical notation and fewer formatting artifacts. The LightOnOCR-2-1B mixture combines teacher-annotated document pages from multiple permissibly usable sources, including scanned material for robustness, as well as auxiliary data to broaden layout coverage.

In addition to full pages, we include document-region crops (paragraphs, headers, abstracts, and general snippets) annotated with \texttt{GPT-4o}~\cite{openai2024gpt4o} to expose the model to varied formats, and we add explicit blank-page examples to enforce a consistent target for empty inputs and mitigate looping or hallucination behaviors. To better cover scientific documents, we also incorporate \TeX{}-derived supervision obtained by compiling raw arXiv sources with an \texttt{nvpdftex}~\cite{nvtexlive2025} pipeline (Section~\ref{nvpdftex}), and we complement the mix with publicly available OCR datasets for additional diversity~\cite{poznanski2025olmocr2}. We release the PDFA-derived annotated subset under license terms that match the underlying PDFA source data as \texttt{\href{https://huggingface.co/datasets/lightonai/LightOnOCR-mix-0126}{lightonai/LightOnOCR-mix-0126}}.

During large-scale annotation, we observed that the teacher occasionally emitted figure bounding box coordinates even when not explicitly prompted to do so. Manual inspection of a sampled subset showed these boxes to be highly accurate. To keep the base OCR objective strictly focused on transcription, we removed coordinate traces from the main supervised training targets; however, we retained them as a separate supervision signal for the bounding-box addition procedure described in Section~\ref{sec:ocr_rlvr}. We also reformatted the coordinates to match our target output convention before training. We publicly release this extracted and normalized subset as \texttt{\href{https://huggingface.co/datasets/lightonai/LightOnOCR-bbox-mix-0126}{lightonai/LightOnOCR-bbox-mix-0126}}.

\subsection{Normalization Pipeline and data clean-up}
\label{sec:normalization}
As described above, our training corpus is assembled from heterogeneous sources (PDFA, scans, open mixtures, arXiv renders) and multiple VLM teachers, which introduces systematic but superficial inconsistencies in the raw transcriptions: stray Markdown code fences, watermark text, variable image placeholders, templated ``this page is empty'' messages, and occasional format drift (e.g., \LaTeX{} environments or HTML fragments where Markdown is expected). While these artifacts rarely affect human readability, they significantly increase target entropy and hurt both deduplication and learning stability. We therefore apply a unified normalization pipeline that maps all sources to a single, canonical target format before mixing and training.

Concretely, normalization is implemented as a sequence of lightweight, mostly deterministic transforms applied prior to hashing and filtering. We (i) sanitize the text (e.g., remove spurious Markdown ticks/code blocks and harmonize whitespace), and apply source-specific cleanup when needed. Notably, we remove recurring watermark artifacts during preprocessing so they do not pollute the deduplication procedure. (ii) homogenize special cases such as full-page embedded images and blank pages by mapping them to fixed targets (respectively a single standard image placeholder and the empty string), and (iii) perform loop and repetition filtering and deduplication on the normalized text. To retain control over the proportion of degenerate-but-common cases, we also separate dedicated pools of empty pages, which were then re-injected at a chosen rate during training as mentioned in Section~\ref{annotations}.

Finally, we run a \LaTeX{} conversion and validation pass to enforce formatting invariants across the full mixture: \LaTeX{} commands are restricted to math spans, headers and sections are converted to markdown, tables are standardized (we use HTML targets to reduce ambiguity), and math expressions are checked for KaTeX~\cite{katex} compatibility. The conversion step emits structured metadata (success/partial/timeout, unresolved references, missing figure numbering, KaTeX compatibility), enabling simple, reproducible filtering rules when constructing the final training mixture.

\subsection{\texttt{nvpdftex} data curation pipeline}
\label{nvpdftex}
High-quality OCR supervision requires pairing rendered page images with faithful, layout-consistent transcriptions. In earlier iterations we relied on a Nougat-style PDF parsing pipeline \cite{Nougat}, but in practice we found it difficult to obtain sufficiently reliable (image, markup) pairs at scale for training: conversion errors, reading order issues, and imperfect alignment between rendered pages and extracted markup introduce noise that can hinder learning. For LightOnOCR-2, we revamped the arXiv extraction pipeline around \texttt{nvpdftex}, a recently released NVIDIA toolchain that hooks directly into the \texttt{pdf\LaTeX{}} engine to produce pixel-aligned annotations without heuristic matching. Concretely, \texttt{nvpdftex} compiles \TeX{} sources and emits (i) a PNG rendering of each page, (ii) structured text targets (Markdown or HTML) for each region, (iii) pixel-accurate bounding boxes with semantic classes (e.g. headers/footers, captions, tables, formulas, pictures), and (iv) page-level metadata (e.g. image dimensions) \cite{nvtexlive2025,eclair2025}. We additionally use this pipeline to generate high-quality figure/image boxes for the arXiv subset used in our bbox RLVR experiments (Section ~\ref{sec:bbox_rlvr}); these arXiv bounding boxes are used for RL supervision rather than included in the main pretraining mixture.

\subsection{LightOnOCR-bbox-bench: New Image Localization Benchmark}
\label{sec:bbox_bench}

\paragraph{LightOnOCR-bbox-bench}
While OCR benchmarks are widely used, there are no standardized benchmarks that specifically measure how well end-to-end vision-language models \emph{localize} images within documents. To fill this gap, we introduce \textbf{LightOnOCR-bbox-bench}, released with this work and composed of two subsets: (i) a manually reviewed subset derived from OlmOCR-Bench~\cite{poznanski2025olmocr} pages (290 samples) and (ii) an automatically annotated arXiv subset generated with \texttt{nvpdftex} (565 samples) and filtered programmatically. We report $F_1$ at IoU threshold 0.5, mean IoU, and count accuracy (exact match on the number of predicted boxes), and evaluate separately on the manual OlmOCR-derived subset and the automatic arXiv+\texttt{nvpdftex} subset.

\section{Training}
\subsection{Pretraining}
\paragraph{Training setup}
We train on a large mixed corpus of OCR datasets described in  Section~\ref{sec:data}, with filtering to remove missing references, numberless figures, KaTeX-incompatible samples, and long completions (capped at 3100 tokens). Document pages are rendered at 200\,DPI during training (except for crops which were already available as images) and resized to a maximum longest edge of 1540 pixels for computational efficiency. We apply moderate document augmentations such as bitmap corruption, erosion/dilation, small affine shears, shift--scale--rotate, 90$^\circ$ rotations, and mild grid distortions, so that a page receives at least one augmentation with probability $0.22$ (and we use a more aggressive regime for blank-page examples). We optimize a next-token prediction objective but mask part of the loss: the loss is computed only on assistant tokens, excluding prompt tokens (system prompt and special tokens) as well as image tokens. Optimization uses AdamW with no weight decay, a peak learning rate of $10^{-4}$, a cosine learning rate schedule with a 100-step warmup, a global batch size of 384 samples, and a maximum sequence length of 6144 tokens. Training is distributed with DDP over $96$ NVIDIA H100 GPUs (80\,GB) using bf16 precision and FlashAttention-2~\cite{dao2023flashattention2fasterattentionbetter}.

\subsection{Reinforcement Learning with Verifiable Rewards}
We apply RLVR~\cite{Lambert2024TLU3P}, where rewards are computed by automatic checks that can be evaluated deterministically on model outputs (e.g., binary unit tests on synthetic documents). This allows us to directly optimize for specific OCR failure modes without having to annotate extra data.

\paragraph{Training setup}
We start from the pretraining checkpoints (\texttt{LightOnOCR-2-1B-base} for OCR and  \texttt{LightOnOCR-2-1B-bbox-base} for localization) and train with GRPO~\cite{shao2024deepseekmath} for one epoch with AdamW at learning rate $4\times10^{-5}$ and KL regularization strength $\beta=0.01$.
We use group-scaled rewards with token-level importance sampling and sample multiple rollouts per prompt (28 for OCR and 14 for bbox). Training is implemented with the Hugging Face TRL library~\cite{huggingface_trl}, and rollouts are generated with vLLM for increased efficiency.

We apply two RLVR recipes: an OCR-focused variant that extends OlmOCR unit tests with additional rewards, and a bbox-focused variant that optimizes IoU-based localization rewards.

\label{sec:rl}
\subsubsection{OlmOCR-2 Style RLVR Recipe}
\label{sec:ocr_rlvr}
We build on the OlmOCR-2 RLVR recipe~\cite{poznanski2025olmocr2}, where rewards are derived from synthetic OlmOCR-Bench–style unit tests, and extend it with additional checks tailored to scientific documents. In addition to scoring completions by the fraction of tests they pass, we (i) penalize low-entropy repetition loops using a compression-based heuristic and ensuring proper EOS termination, (ii) reward mathematical correctness by extracting math spans and verifying they render successfully with KaTeX, (iii) enforce clean math formatting by rejecting common artifacts such as HTML tags, misused Markdown italics for variables, and unbalanced \LaTeX{} delimiters or environments, and (iv) remove the frontmatter reward that encourages the model to output document metadata at the top of its response. Finally, for the headers/footers category, we change the \emph{absence} reward to instead reward \emph{presence} of headers, footers, and page numbers, encouraging high-fidelity extraction of all visible content.

\subsubsection{Bounding Box RLVR Recipe}
\label{sec:bbox_rlvr}
To complement OCR with image localization, we extend the output format for embedded images to
\texttt{![image](image\_N.png)x1,y1,x2,y2}, where coordinates are normalized to $[0,1000]$. Throughout pretraining, the model is already trained to emit the image placeholder \texttt{![image](image\_N.png)} whenever an image is present; bounding box learning therefore mainly adds \emph{where} the image is, not \emph{whether} it exists. We enable this capability by resuming pretraining from the base checkpoint and continuing on a mixture stage that includes bbox-annotated pages, providing a cold start for RL. This pretraining integration preserves OCR performance, enabling joint OCR and image localization.

We then further improve image localization with RLVR using an IoU-based reward The full reward definition is given in Appendix~\ref{app:bbox_reward}.

\subsubsection{Model Averaging and Merging}
We use \textbf{checkpoint averaging} by souping the last $5$ checkpoints, which consistently outperforms any single checkpoint. This yields our strong supervised pretraining baseline \texttt{LightOnOCR-2-1B-base}. We then apply \textbf{RLVR} on top of this base to obtain our best OCR model \texttt{LightOnOCR-2-1B}. In parallel, we train a bbox-capable variant by resuming supervised pretraining with coordinate annotations introduced during training, producing \texttt{LightOnOCR-2-1B-bbox-base}, and further refine it with bbox-focused RLVR to obtain \texttt{LightOnOCR-2-1B-bbox}. Finally, we apply \textbf{task arithmetic} merging,
$\theta_{\text{merge}}=\theta_{\text{base}}+\alpha(\theta_{\text{rl}}-\theta_{\text{base}})$,
to form \texttt{LightOnOCR-2-1B-ocr-soup} by merging \texttt{LightOnOCR-2-1B} into the averaged base with $\alpha=0.4$. We then construct \texttt{LightOnOCR-2-1B-bbox-soup} by merging \texttt{LightOnOCR-2-1B-ocr-soup} into the bbox-specialized checkpoint \texttt{LightOnOCR-2-1B-bbox} (used as the base) with $\alpha=0.1$, yielding an explicit OCR--bbox trade-off without additional training. Our training targets retain headers/footers, and in RLVR we flip OlmOCR's header/footer \emph{absence} tests to reward \emph{presence} of this text. We also applied the OCR RLVR recipe to \texttt{LightOnOCR-1B-1025} to obtain \texttt{LightOnOCR-1B-1025-GRPO} for reference.

\section{Results}
\label{sec:results}

We evaluate LightOnOCR-2-1B on OlmOCR-Bench~\cite{poznanski2025olmocr} as our primary OCR benchmark. We additionally report OmniDocBench v1.0~\cite{omnidocbench} results in Appendix~\ref{app:omnidocbench}. For localization, we evaluate on LightOnOCR-bbox-bench (Section~\ref{sec:bbox_bench}). Unless otherwise stated, we use a single-pass evaluation without test-time heuristics (e.g., rotation sweeps or retries); inference details for LightOnOCR models are provided in Appendix~\ref{app:inference_details}.

\subsection{OlmOCR-Bench}

\begin{table}[t]
\centering
\small
\setlength{\tabcolsep}{2pt}
\begin{adjustbox}{max width=\linewidth, keepaspectratio}
\begin{tabular}{lcccccccccc}
\toprule
\textbf{Model} & \textbf{End-to-end} & \textbf{Size (B)} & \textbf{ArXiv} & \textbf{Old Scans Math} & \textbf{Tables} & \textbf{Old Scans} & \textbf{Multi-column} & \textbf{Long Tiny Text} & \textbf{Base} & \textbf{Overall} \\
\midrule
% Marker v1.7.5 & $\times$ &  & 76.0 & 57.9 & 57.6 & 27.8 & 72.9 & 84.6 & 99.1 & 68.0 $\pm$ 1.1 \\
% MinerU v1.3.10 & $\times$ &  & 75.4 & 47.4 & 60.9 & 17.3 & 59.0 & 39.1 & 96.6 & 56.5 $\pm$ 1.1 \\
% Mistral OCR API & - &  & 77.2 & 67.5 & 60.6 & 29.3 & 71.3 & 77.1 & 99.4 & 68.9 $\pm$ 1.1 \\
Mistral OCR 3 API & - & - & 85.6 & 69.7 & 85.5 & 43.5 & 81.2 & 88.5 & 99.7 & 79.1 $\pm$ 1.0 \\
% GPT-4o & - &  & 51.5 & 75.5 & 69.1 & 40.9 & 68.9 & 54.1 & 96.7 & 65.2 $\pm$ 1.1 \\
Gemini Flash 2 & - & - & 32.1 & 56.3 & 61.4 & 27.8 & 58.7 & 84.4 & 94.0 & 59.2 $\pm$ 1.1 \\
% \multicolumn{10}{l}{\textit{$\geq 2.0\textrm{B}$ Parameters}}\\
% Nanonets-OCR-s & $\times$ & 3.75 & 67.0 & 68.6 & 77.7 & 39.5 & 69.9 & 53.4 & 99.3 & 67.9 $\pm$ 1.1 \\
% Qwen 2 VL & $\checkmark$ & 8 & 19.7 & 31.7 & 24.2 & 17.1 & 8.3 & 6.8 & 55.5 & 23.3 $\pm$ 0.9 \\
Qwen2.5-VL-8B & $\checkmark$ & 8 & 63.1 & 65.7 & 67.3 & 38.6 & 68.3 & 49.1 & 98.3 & 64.3 $\pm$ 1.2 \\
olmOCR v0.3.0 & $\times$ & 8 & 78.6 & 79.9 & 72.9 & 43.9 & 77.3 & 81.2 & 98.9 & 76.1 $\pm$ 1.1 \\
MonkeyOCR-pro-3B & $\times$ & 3 & 83.8 & 68.8 & 74.7 & 36.1 & 76.6 & 80.1 & 95.3 & 73.6 $\pm$ 1.0 \\
dots.ocr & $\times$ & 3 & 82.1 & 64.2 & 88.3 & 40.9 & 82.4 & 81.2 & 99.5 & 76.9 $\pm$ 1.0 \\
dots.ocr-1.5 & $\times$ & 3 & 85.9 & 85.5 & 90.7 & 48.2 & 85.3 & 81.6 & 99.7 & 82.4 $\pm$ 0.9 \\
DeepSeek-OCR & $\checkmark$ & 3 & 77.5 & 74.5 & 77.3 & 33.1 & 67.3 & 83.0 & 99.3 & 73.1 $\pm$ 1.0 \\
% DeepSeek-OCR-2 & $\checkmark$ & 3 & 82.0 & 72.0 & 77.4 & 33.8 & 79.0 & 90.7 & 99.6 & 76.3 $\pm$ 0.9 \\
Chandra-9B & $\checkmark$ & 9 & 82.2 & 80.3 & 88.0 & \bestA{50.4} & 81.2 & \bestA{92.3} & \bestA{99.9} & 81.7 $\pm$ 0.9 \\
olmOCR-2-8B & $\checkmark$ & 8 & 82.9 & \textbf{82.1} & 84.3 & \textbf{48.3} & 84.3 & 81.4 & 99.7 & 80.4 $\pm$ 1.1 \\
% \multicolumn{10}{l}{\textit{$\leq 2.0\textrm{B}$ Parameters}}\\
% GOT OCR & $\checkmark$ & 0.56 & 52.7 & 52.0 & 0.20 & 22.1 & 42.0 & 29.9 & 94.0 & 41.8 $\pm$ 1.1 \\
MonkeyOCR-pro-1.2B & $\times$ & 1.2 & 80.5 & 62.9 & 71.1 & 32.9 & 68.3 & 74.0 & 92.6 & 68.9 $\pm$ 1.1 \\
MinerU2.5 & $\times$ & 1.2 & 76.6 & 54.6 & 84.9 & 33.7 & 78.2 & 81.2 & 83.5 & 70.4 $\pm$ 1.0 \\
PaddleOCR-VL & $\times$ & 0.9 & 85.7 & 71.0 & 84.1 & 37.8 & 79.9 & 85.7 & 98.5 & 77.5 $\pm$ 1.0 \\
% Qwen3-VL-2B-Instruct & $\checkmark$ & 2 & 67.5 & 67.2 & 61.6 & 28.1 & 54.1 & 44.6 & 97.8 & 60.1 $\pm$ 1.1 \\

\midrule
LightOnOCR-1B-1025 & $\checkmark$ & 1 & 81.4 & 71.6 & 76.4 & 35.2 & 80.0 & 88.7 & 99.5 & 76.1 $\pm$ 1.1 \\
LightOnOCR-1B-1025-GRPO & $\checkmark$ & 1 & 86.5 & 73.8 & 74.5 & 32.9 & 85.1 & \textbf{91.6} & 99.7 & 77.7 $\pm$ 1.0 \\
\midrule
% LightOnOCR-2-1B-base(last5 avg) & $\checkmark$ & 1 & \bestB{84.6} & 80.5 & 87.4 & 46.5 & 84.3 & 89.1 & 99.6 & 81.7 \\

LightOnOCR-2-1B-base & $\checkmark$ & 1 & 84.9 & 80.3 & 86.7 & 47.0 & 84.6 & 89.1 & 99.8 & 81.8 $\pm$ 0.9  \\

LightOnOCR-2-1B-bbox-base & $\checkmark$ & 1 & 84.6 & 78.6 & 84.7 & 46.0 &  83.8 & 88.0 & 99.8 & 80.8 $\pm$ 0.9 \\

LightOnOCR-2-1B-bbox & $\checkmark$ & 1 & \textbf{86.9} & 74.7 & 88.6 & 39.7 & \textbf{85.0} & 86.4 & 99.8 & 80.2 $\pm$ 0.9\\

LightOnOCR-2-1B-bbox-soup & $\checkmark$ & 1 & 86.1 & 77.9 & 88.2 & 41.2 & \bestA{85.4} & 87.3 & 99.7 & 80.8 $\pm$ 0.9\\

LightOnOCR-2-1B-ocr-soup & $\checkmark$ & 1 & 86.8 & 81.2 & \bestA{89.0} & 45.4 &  84.2 & 90.3 & 99.7 & \textbf{82.4 $\pm$ 0.9} \\

\textbf{LightOnOCR-2-1B} & $\checkmark$  & 1 & \bestA{89.6} & \bestA{85.6} & \bestA{89.0} & 42.2 & 84.8 & 91.4 & 99.6 & \bestA{83.2 $\pm$ 0.9} \\

\bottomrule
\\
\end{tabular}
\end{adjustbox}
\caption{
OlmOCR-Bench results (headers/footers category excluded; see Appendix~\ref{app:headers_footers}).
Per-column best is highlighted in blue and second best in bold. Results are taken from the corresponding published works; we additionally evaluate DeepSeekOCR and the Mistral OCR 3 API since they do not report OlmOCR-Bench numbers.
Inference details for our models are provided in Appendix~\ref{app:inference_details}.
}
\label{tab:olmocr_bench_main}
\end{table}

We report OlmOCR-Bench results without the \emph{headers/footers} category as originally defined, since it rewards \emph{omitting} visible content (e.g., titles and page numbers) whereas our objective is full-page transcription; systems that explicitly suppress these regions can score well without improving full-page extraction quality, and even empty outputs can achieve a perfect score. We evaluate all models without test-time heuristics (e.g., retries or rotation correction) to reflect raw model behavior.

Table~\ref{tab:olmocr_bench_main} shows that \texttt{LightOnOCR-2-1B} achieves the highest overall score ($83.2 \pm 0.9$) among evaluated systems, outperforming substantially larger end-to-end models while using only 1B parameters and being end-to-end trainable. Improvements are broad across categories, with particularly strong scores on ArXiv, old scans math, and table-heavy documents, highlighting the benefit of higher-quality data, increased scientific coverage, and higher-resolution training. 

Comparing \texttt{LightOnOCR-2-1B-base} to \texttt{LightOnOCR-2-1B} isolates the effect of RLVR, which improves overall performance and reduces common generation failures such as repetition loops (Appendix~\ref{app:repetitions}). Introducing bbox prediction (\texttt{LightOnOCR-2-1B-bbox}) yields a small OCR quality drop relative to the base checkpoint while enabling localization; task-arithmetic merging partially recover OCR performance and provide an explicit OCR--bbox trade-off (\texttt{bbox-soup}). As we optimize for the presence of headers/footers, under the original OlmOCR-Bench definition (rewarding suppression), our \texttt{headers\_footers} score decreases; see Appendix~\ref{app:headers_footers} for details. When applying the same recipe to our previous model, we observe similar gains on OCR performance, see \texttt{LightOnOCR-1B-1025-GRPO} in Table~\ref{tab:olmocr_bench_main}.

% \paragraph{Repetitions}
% We provide an auxiliary analysis of repetition loops and their reduction under RLVR in Appendix~\ref{app:repetitions}.

\subsection{Image Bounding Box Detection}
\label{sec:results_bbox_no_regression}
Table~\ref{tab:bbox_results} reports bounding-box detection on LightOnOCR-bbox-bench. The goal is to measure whether a compact end-to-end OCR model can localize embedded images while retaining strong transcription quality. We compare against Chandra-9B~\cite{chandra_repo} under the same evaluation protocol and restrict evaluation to visual elements (figures/images), ignoring other layout categories.

\begin{table}[htbp!]
\centering
\caption{Bounding box detection on LightOnOCR-bbox-bench.}
\label{tab:bbox_results}
\small
\setlength{\tabcolsep}{6pt}
\begin{tabular}{l ccc ccc}
\toprule
& \multicolumn{3}{c}{\textbf{OlmOCR (290)}} & \multicolumn{3}{c}{\textbf{arXiv (565)}} \\
\cmidrule(lr){2-4}\cmidrule(lr){5-7}
\textbf{Model} & \textbf{$F_1@0.5$} & \textbf{IoU} & \textbf{Count Acc.} & \textbf{$F_1@0.5$} & \textbf{IoU} & \textbf{Count Acc.} \\
\midrule
Chandra-9B & 0.75 & \textbf{0.71} & 75.2 & 0.81 & \textbf{0.77} & 81.8 \\
\midrule
LightOnOCR-2-1B-bbox & \textbf{0.78} & 0.70 & \textbf{83.8} & \textbf{0.83} & \textbf{0.77} & \textbf{85.0} \\
LightOnOCR-2-1B-bbox-soup & 0.76 & 0.67 & 80.7 & 0.82 & 0.76 & 85.1 \\

% LightOnOCR-2-1B-bbox(bbox-soup, 10k samples) & 0.698 & 0.608 & 75.9\% & 0.811 & 0.740 & 84.2\% \\
% \midrule
\bottomrule
\end{tabular}
\end{table}

\paragraph{Localization quality}
\texttt{LightOnOCR-2-1B-bbox} improves $F_1@0.5$ and count accuracy over the 9B baseline on both subsets, while achieving comparable mean IoU. This indicates reliable detection of both the presence and number of figures, with accurate localization, despite the substantially smaller model size.

\subsection{Efficiency}
\label{sec:efficiency}
Beyond accuracy, we measure throughput to characterize the practical speed--quality trade-off of end-to-end OCR models. We measure inference efficiency by running the full OlmOCR-Bench evaluation (1,403 pages) end-to-end and reporting pages/sec as the total number of pages divided by the wall-clock time to complete the benchmark. We prefer pages/sec over tokens/sec, since tokenization and output lengths differ across models and formats, making tokens/sec less comparable.
Each model was run using its official library and the inference parameters recommended by its respective authors to ensure a fair comparison. Table~\ref{tab:efficiency_pages_per_sec} compares LightOnOCR-2-1B against the main end-to-end baselines; a full comparison with additional systems is provided in Appendix~\ref{app:efficiency_full}.

\begin{table}[htbp!]
\centering
\small
\setlength{\tabcolsep}{6pt}
\begin{tabular}{lccc}
\toprule
\textbf{Model} & \textbf{Dtype} & \textbf{Size (B)} & \textbf{Throughput (pages/sec)} \\
\midrule
LightOnOCR-2 & BF16 & 1 & \textbf{5.71} \\
olmOCR-2 & FP8 & 8 & 3.28 \\
Chandra & BF16 & 9 & 1.70 \\
\bottomrule
\end{tabular}
\caption{Inference throughput on a single NVIDIA H100 (80\,GB).}
\label{tab:efficiency_pages_per_sec}
\end{table}

These results show that LightOnOCR-2 provides substantially higher throughput than larger end-to-end baselines, making it practical for high-volume document processing.

\section{Scope and Limitations}
\label{sec:scope_limitations}

LightOnOCR-2-1B models are designed for printed document understanding and perform particularly well on:
(i) scientific PDFs, including dense typography and accurate \LaTeX{} math transcription (e.g., strong arXiv and ``old scans math'' results in Table~\ref{tab:olmocr_bench_main});
(ii) scans of typed documents, including moderately degraded, noisy or rotated scans;
(iii) European languages and Latin scripts, reflecting the distributional emphasis of the pretraining mixture;
and (iv) layout-heavy pages such as multi-column documents and long-form tables, where faithful reading order and structure are critical.

Despite strong overall robustness, we note two important limitations.
First, multilingual performance outside of European / Latin-script languages is not fully supported: our training mix and normalization pipeline prioritize Latin-script documents, and some non-Latin scripts (e.g., CJK or Arabic) can exhibit degraded fidelity or inefficient tokenization compared to the in-scope languages (as shown in Appendix~\ref{sec:vocab_pruning} it is as expected all the more evident for pruned models).
Second, handwritten text transcription remains inconsistent: while LightOnOCR-2-1B benefits from scan coverage, its supervision is primarily derived from printed or typeset sources, and cursive or unconstrained handwriting is not a target use-case for the released checkpoints.
We view these as promising directions for future work through targeted data collection and evaluation.

\section{Conclusion}
We introduce LightOnOCR-2-1B a 1B-parameter end-to-end OCR VLM setting new state-of-the-art on OlmoOCR-Bench. Building on top of the LightOnOCR architecture, setup and insight, we detailed the key factors behind its improved performance: a substantially larger and cleaner pretraining mixture, stronger document normalization and conversion pipelines, and higher-resolution training with targeted augmentations. We further presented an image-localization variant trained to predict bounding boxes without degrading OCR by introducing coordinates during pretraining and refining them with RLVR. Finally, we showed that lightweight weight-space techniques, checkpoint averaging and task-arithmetic merging, provide practical gains and enable explicit control of the OCR--bbox trade-off. We release model weights, datasets, and the LightOnOCR-bbox-bench benchmark to support reproducible research on high-fidelity document extraction and localization.

\section*{Acknowledgments}

This work was granted access to the HPC resources of IDRIS under GENCI allocations \texttt{AS011016449}, \texttt{A0181016214}, and \texttt{A0171015706}, enabling us to use the Jean Zay supercomputer. All RL experiments were run on our in-house H200 nodes. We thank Stéphane Réquena and the IDRIS support team for their valuable help. We also thank the LightOn team for their support in making this release possible, with special thanks to Antoine Chaffin and Amélie Chatelain for their help throughout the release process.

The project received funding from the BPI Scribe project.

\bibliographystyle{unsrt}
\bibliography{references}
\newpage
\appendix
\section{Vocabulary Pruning}
\label{sec:vocab_pruning}

The Qwen3 decoder uses a 151,936-token multilingual vocabulary, much of which is unused for language-specific OCR. We investigate frequency-based vocabulary pruning for English/French documents, reducing to 51k, 32k, and 16k tokens while preserving tokenizer integrity through recursive sub-token frequency propagation.

Table~\ref{tab:vocab_pruning} summarizes the trade-offs. Pruning to 16k tokens reduces parameters by 13.8\% with minimal OCR degradation on English benchmarks (75.4\% vs 76.1\% on OlmOCR-Bench). The 32k variant achieves the best speed-accuracy balance: 11.6\% faster inference while retaining 96\% of base performance. However, non-Latin scripts (Arabic, Chinese) experience ${\sim}3\times$ token count inflation as script-specific tokens are removed. These experiments were conducted on LightOnOCR-1; we release the pruned variants as \texttt{LightOnOCR-0.9B-32k-1025}\footnote{\url{https://huggingface.co/lightonai/LightOnOCR-0.9B-32k-1025}} and \texttt{LightOnOCR-0.9B-16k-1025}\footnote{\url{https://huggingface.co/lightonai/LightOnOCR-0.9B-16k-1025}}.

\begin{table}[h]
\centering
\small
\setlength{\tabcolsep}{4pt}
\begin{tabular}{lcccc}
\toprule
\textbf{Vocab} & \textbf{Params (M)} & \textbf{OlmOCR Overall} & \textbf{Speedup vs Base} & \textbf{Tokens/page (EN/ZH)} \\
\midrule
151k (Base) & 1005.6 & 76.1 & -- & 475 / 950 \\
51k & 902.5 & 67.7 & $+9.5\%$ & 485 / 2220 \\
32k & 883.6 & 73.1 & $+11.6\%$ & 510 / 2750 \\
16k & 866.8 & 75.4 & $+3.9\%$ & 575 / 3200 \\
\bottomrule
\end{tabular}
\caption{Vocabulary pruning trade-offs. Smaller vocabularies reduce parameters and improve speed for Latin-script languages but degrade tokenization for Chinese (ZH) and other non-Latin scripts.}
\label{tab:vocab_pruning}
\end{table}

\section{Inference Details}
\label{app:inference_details}
LightOnOCR models are evaluated with $T=0.2$, $\mathrm{top}\_p=0.9$ and $\mathrm{top}\_k=0$.
For OlmOCR-Bench, we use three independent generations(3 repeats) per page; maximum resolution set to 1540, no test-time heuristics (retries/rotations) are used.

\section{Additional Results}
\subsection{OlmOCR Headers/Footers}
\label{app:headers_footers}
OlmOCR-Bench includes a \texttt{headers\_footers} category whose unit tests reward \emph{absence} of header and footer text. In contrast, our training objective is full-page transcription: throughout both pretraining and RLVR, our targets retain all visible content, including page numbers, running headers, titles, and footers. Moreover, in RLVR we explicitly flip the header/footer \emph{absence} tests to reward \emph{presence} of header and footer text. As a result, when evaluated under the \emph{original} OlmOCR-Bench scoring, models optimized for full-page extraction can obtain lower \texttt{headers\_footers} scores. Table~\ref{tab:headers_footers_effect} reports this metric for all released checkpoints.

\begin{table}[h]
\centering
\small
\setlength{\tabcolsep}{6pt}
\begin{tabular}{lc}
\toprule
\textbf{Checkpoint} & \textbf{OlmOCR \texttt{headers\_footers} score} \\
\midrule
LightOnOCR-2-1B            & \bestA{19.74} \\
LightOnOCR-2-1B-base       & 31.05 \\
LightOnOCR-2-1B-bbox       & 29.34 \\
LightOnOCR-2-1B-bbox-base  & 31.05 \\
LightOnOCR-2-1B-bbox-soup  & 29.47 \\
LightOnOCR-2-1B-ocr-soup   & 25.13 \\
\bottomrule
\end{tabular}
\caption{OlmOCR-Bench \texttt{headers\_footers} scores under the \emph{original} benchmark definition (rewarding absence of header/footer text). Our models are trained for full-page transcription, so this metric is not aligned with our objective.}
\label{tab:headers_footers_effect}
\end{table}

\subsection{OmniDocBench Results}
\label{app:omnidocbench}
We report results on OmniDocBench v1.0~\cite{omnidocbench} in Table~\ref{omnidocbench}. Its language-split reporting provides a complementary view—most notably for reading order and structured layout fidelity. However, OmniDocBench relies heavily on edit-distance-based metrics that are sensitive to formatting conventions, and the benchmark primarily targets English and Chinese documents. We therefore treat it as a secondary evaluation signal.

\begin{table}[h!]
\centering
\small
\setlength{\tabcolsep}{4pt}
\begin{adjustbox}{max width=\linewidth, keepaspectratio}
\begin{tabular}{l c c c c c c c c c c c c c c c}
\toprule
\multirow{2}{*}{\textbf{Model}} & \multirow{2}{*}{\textbf{Size (B)}} & \multicolumn{2}{c}{\textbf{Overall}$^{\text{Edit}}$↓} & \multicolumn{2}{c}{\textbf{Text}$^{\text{Edit}}$↓} & \multicolumn{2}{c}{\textbf{Formula}$^{\text{Edit}}$↓} & \multicolumn{2}{c}{\textbf{Formula}$^{\text{CDM}}$↑} & \multicolumn{2}{c}{\textbf{Table}$^{\text{TEDS}}$↑} & \multicolumn{2}{c}{\textbf{Table}$^{\text{Edit}}$↓} & \multicolumn{2}{c}{\textbf{Read Order}$^{\text{Edit}}$↓} \\
\cmidrule(lr){3-16}
& & EN & ZH & EN & ZH & EN & ZH & EN & ZH & EN & ZH & EN & ZH & EN & ZH \\
\midrule
% Marker-1.7.1 &  & 0.296 & 0.497 & 0.085 & 0.293 & 0.374 & 0.688 & 79.0 & 36.7 & 67.6 & 54.0 & 0.609 & 0.678 & 0.116 & 0.329 \\
% Mistral OCR &  & 0.268 & 0.439 & 0.072 & 0.325 & 0.318 & 0.495 & 64.6 & 45.9 & 75.8 & 63.6 & 0.6 & 0.65 & 0.083 & 0.284 \\
% GPT4o &  & 0.233 & 0.399 & 0.144 & 0.409 & 0.425 & 0.606 & 72.8 & 42.8 & 72.0 & 62.9 & 0.234 & 0.329 & 0.128 & 0.251 \\
Gemini2.0-flash & - & 0.191 & 0.264 & 0.091 & 0.139 & 0.389 & 0.584 & 77.6 & 43.6 & 79.7 & 78.9 & 0.193 & 0.206 & 0.092 & 0.128 \\
% Nanonets-OCR-s & 3.75 & 0.283 & 0.295 & 0.134 & 0.231 & 0.518 & 0.546 & 63.2 & 52.0 & 76.8 & 79.4 & 0.343 & 0.201 & 0.135 & 0.2 \\
Qwen2-VL-72B & 72 & 0.252 & 0.327 & 0.096 & 0.218 & 0.404 & 0.487 & \bestA{82.2} & \bestA{61.2} & 76.8 & 76.4 & 0.387 & 0.408 & 0.119 & 0.193 \\
% Qwen2.5-VL-7B & 8 & 0.316 & 0.399 & 0.151 & 0.243 & 0.376 & 0.5 & 75.3 & 57.3 & 71.1 & 71.3 & 0.598 & 0.627 & 0.138 & 0.226 \\
% OLMOCR-sglang & 8 & 0.326 & 0.469 & 0.097 & 0.293 & 0.455 & 0.655 & 74.3 & 43.2 & 68.1 & 61.3 & 0.608 & 0.652 & 0.145 & 0.277 \\
MonkeyOCR-pro-3B & 3 & 0.138 & 0.206 & 0.067 & 0.107 & 0.246 & 0.421 &  &  & 81.5 & 87.5 & 0.139 & 0.111 & 0.100 & 0.185 \\
dots.ocr & 3 & 0.125 & 0.160 & \bestA{0.032} & 0.066 & 0.329 & 0.416 &  &  & \bestA{88.6} & 89.0 & 0.099 & 0.092 & \bestA{0.040} & 0.067 \\
DeepSeek-OCR (Gundam-M) & 3 & 0.123 & 0.157 & 0.049 & 0.087 & 0.242 & 0.377 &  &  &  &  & 0.147 & 0.08 & 0.056 & 0.085 \\
% GOT-OCR & 0.56 & 0.287 & 0.411 & 0.189 & 0.315 & 0.360 & 0.528 & 74.3 & 45.3 & 53.2 & 47.2 & 0.459 & 0.52 & 0.141 & 0.28 \\
MonkeyOCR-pro-1.2B & 1.2 & 0.146 & 0.221 & 0.068 & 0.118 & 0.272 & 0.452 & 76.663 & 63.282 & 81.342 & 85.504 & 0.149 & 0.134 & 0.093 & 0.179 \\
MinerU2.5 & 1.2 & 0.111 & 0.174 & 0.050 & 0.074 & 0.258 & 0.473 &  &  & 88.3 & 89.2 & 0.089 & 0.083 & 0.045 & 0.068 \\
PaddleOCR-VL & 0.9 & \bestA{0.105} & \bestA{0.126} & 0.041 & \bestA{0.062} & \bestA{0.241} & \bestA{0.316} &  &  & 88.0 & \bestA{92.1} & \bestA{0.093} & \bestA{0.062} & 0.045 & \bestA{0.063} \\
\midrule
LightOnOCR-1B-1025 & 1 & 0.234 & 0.392 & 0.090 & 0.380 & 0.486 & 0.621 &  &  & 70.1 & 63.7 & 0.262 & 0.307 & 0.098 & 0.259 \\
LightOnOCR-1B-1025-GRPO & 1 & 0.254 & 0.390 & 0.267 & 0.463 & 0.384 & 0.586 &  &  & 65.3 & 59.6 & 0.305 & 0.316 & 0.061 & 0.195 \\
\midrule
\textbf{LightOnOCR-2-1B} & 1 & 0.146 & 0.263 & 0.082 & 0.317 & 0.339 & 0.438 &  &  & 85.7 & 82.7 & 0.112 & 0.132 & 0.050 & 0.167 \\
LightOnOCR-2-1B-bbox & 1 & 0.152 & 0.305 & 0.080 & 0.343 & 0.316 & 0.523 &  &  & 82.1 & 78.0 & 0.144 & 0.171 & 0.068 & 0.181 \\
LightOnOCR-2-1B-base & 1 & 0.150 & 0.281 & 0.059 & 0.289 & 0.340 & 0.530 &  &  & 82.9 & 80.5 & 0.140 & 0.144 & 0.060 & 0.161 \\
LightOnOCR-2-1B-bbox-base & 1 & 0.153 & 0.277 & 0.062 & 0.303 & 0.341 & 0.489 &  &  & 82.0 & 80.9 & 0.146 & 0.145 & 0.064 & 0.173 \\
LightOnOCR-2-1B-ocr-soup & 1 & 0.150 & 0.255 & 0.071 & 0.295 & 0.338 & 0.437 &  &  & 84.3 & 83.1 & 0.126 & 0.124 & 0.066 & 0.164 \\
\bottomrule
\\
\end{tabular}
\end{adjustbox}
\caption{OmniDocBench v1.0 (EN/ZH) results. Per-column best within each size group is highlighted in blue. LightOnOCR-2 shows a clear improvement over LightOnOCR-1, with \texttt{LightOnOCR-2-1B} ranking among the strongest models in its size class.}
\label{omnidocbench}
\end{table}

\subsection{Extended Efficiency Comparison}
\label{app:efficiency_full}
Table~\ref{tab:efficiency_full} extends the efficiency comparison from Section~\ref{sec:efficiency} to include additional OCR systems. Each model was run using its official library and the inference parameters recommended by its respective authors. We prefer pages/sec over tokens/sec since tokenization and output lengths differ across models, making tokens/sec less comparable.

\begin{table}[htbp!]
\centering
\small
\setlength{\tabcolsep}{6pt}
\begin{tabular}{lcccc}
\toprule
\textbf{Model} & \textbf{Dtype} & \textbf{Size (B)} & \textbf{Throughput (pages/sec)} & \textbf{Speedup vs slowest} \\
\midrule
LightOnOCR-2 & BF16 & 1 & \textbf{5.71} & 6.49$\times$ \\
olmOCR-2 & FP8 & 8 & 3.28 & 3.73$\times$ \\
olmOCR-2 & BF16 & 8 & 2.54 & 2.89$\times$ \\
DeepSeek-OCR & BF16 & 3 & 2.36 & 2.68$\times$ \\
PaddleOCR-VL & BF16 & 1 & 2.14 & 2.43$\times$ \\
Chandra & BF16 & 9 & 1.70 & 1.93$\times$ \\
dots.ocr & BF16 & 3 & 0.88 & 1.00$\times$ \\
\bottomrule
\end{tabular}
\caption{Full inference throughput comparison on a single NVIDIA H100 (80\,GB).}
\label{tab:efficiency_full}
\end{table}

\subsection{RLVR Effect on Repetition Loops}
\label{app:repetitions}
To quantify repetition loops, we scan OlmOCR-Bench generations with a low-entropy detector based on the ZLIB compression ratio (flagging outputs with ratio $<0.13$). Table~\ref{tab:repetition_loops} summarizes the fraction of flagged generations, showing a reduction after RLVR, consistent with our repetition-focused reward component (Section~\ref{sec:rl}).

\begin{table}[h]
\centering
\small
\setlength{\tabcolsep}{8pt}
\begin{tabular}{lccc}
\toprule
\textbf{Checkpoint} & \textbf{\% Loopy} \\
\midrule
\texttt{LightOnOCR-2-1B-base} & 1.14\% \\
\texttt{LightOnOCR-2-1B} & 0.50\% \\
\bottomrule
\end{tabular}
\caption{Loopy generations on OlmOCR-Bench.}
\label{tab:repetition_loops}
\end{table}

% Appendix
\section{Bounding Box RLVR Reward}
\label{app:bbox_reward}
Let $I_{\text{gt}}$ and $I_{\text{pred}}$ denote the sets of image IDs present in the ground-truth and predicted outputs for a page, and let $\mathcal{B}^{\text{gt}}_i$ and $\mathcal{B}^{\text{pred}}_i$ be the corresponding boxes for image ID $i$. We compute the mean IoU over matched IDs and scale it by an ID-overlap factor:
\begin{equation}
R_{\text{bbox}} \;=\;
\left(\frac{1}{|I_{\cap}|}\sum_{i \in I_{\cap}} \mathrm{IoU}\!\left(\mathcal{B}^{\text{pred}}_i, \mathcal{B}^{\text{gt}}_i\right)\right)
\cdot
\frac{|I_{\cap}|}{\max\!\left(|I_{\text{gt}}|, |I_{\text{pred}}|\right)},
\qquad
I_{\cap}= I_{\text{gt}} \cap I_{\text{pred}} .
\end{equation}
This formulation rewards accurate localization for correctly predicted image IDs while penalizing missing and hallucinated boxes through the overlap factor.

% \section{\texttt{nvpdftex} Failure Modes}
% \label{app:nvpdftex_failures}
% In practice, we encountered several non-obvious failure modes that required explicit cleaning: silent compilation failures that surfaced downstream as \texttt{\textbackslash unknown} commands (about 12\% of samples), invalid geometry such as negative coordinates, and degenerate \TeX{} patterns such as \texttt{\textbackslash includegraphics\{\}} loops (about 8.3\% of samples). We also observe occasional over-segmentation and spurious boxes (e.g., fragmented references or mislocalized regions), which we mitigate through filtering, deduplication, and normalization.

\subsection{Task-Arithmetic Model Merging}
\label{sec:model_merging}
\begin{figure}[h]
    \centering
    \includegraphics[width=0.7\columnwidth]{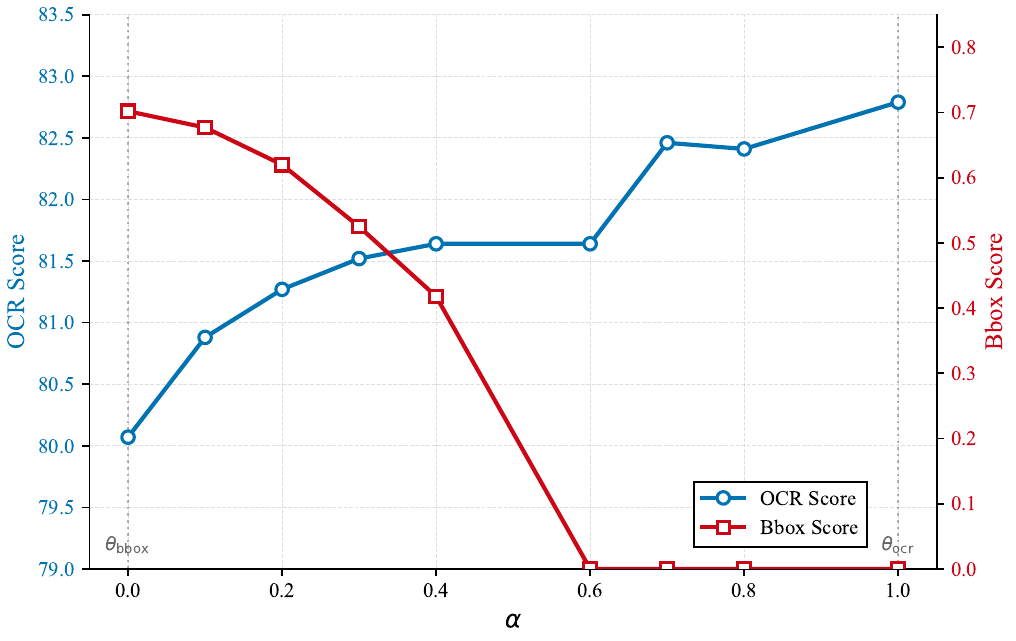}
    \caption{Task-arithmetic interpolation between a bbox-specialized and an OCR-specialized checkpoint.}
    \label{fig:param_merge}
\end{figure}

Because OCR quality and localization accuracy can pull model behavior in different directions, we use weight-space merging to expose a controllable trade-off between the two directions. We apply task-arithmetic merging \cite{ilharco2023taskarithmetic} to combine an OCR-specialized checkpoint with a bounding-box (bbox) specialized checkpoint while explicitly controlling their trade-off. Concretely, we interpolate in weight space as
$\theta = \theta_{\text{bbox}} + \alpha(\theta_{\text{ocr}} - \theta_{\text{bbox}})$ with $\alpha\in[0,1]$, and evaluate the resulting models on both OCR and bbox localization metrics (Figure~\ref{fig:param_merge}). The difference vector $(\theta_{\text{ocr}}-\theta_{\text{bbox}})$ can be interpreted as an OCR ``task vector'' added to the bbox model with strength $\alpha$, yielding a single merged checkpoint with no additional training cost and unchanged inference. As expected, OCR performance increases with $\alpha$, but bbox detection degrades beyond $\alpha>0.4$ and collapses for $\alpha\geq 0.6$, suggesting that localization-specific parameters are progressively overwritten. In our experiments, the best balance occurs around $\alpha\approx 0.1$, which preserves strong bbox quality (IoU$=0.677$) while improving OCR performance to 80.88\%.

\newpage
\section{Examples}
We provide here some examples of transcriptions highlighting the capabilities or the OCR model family. 
\begin{figure}[h!]
    \centering
    \includegraphics[width=1\linewidth]{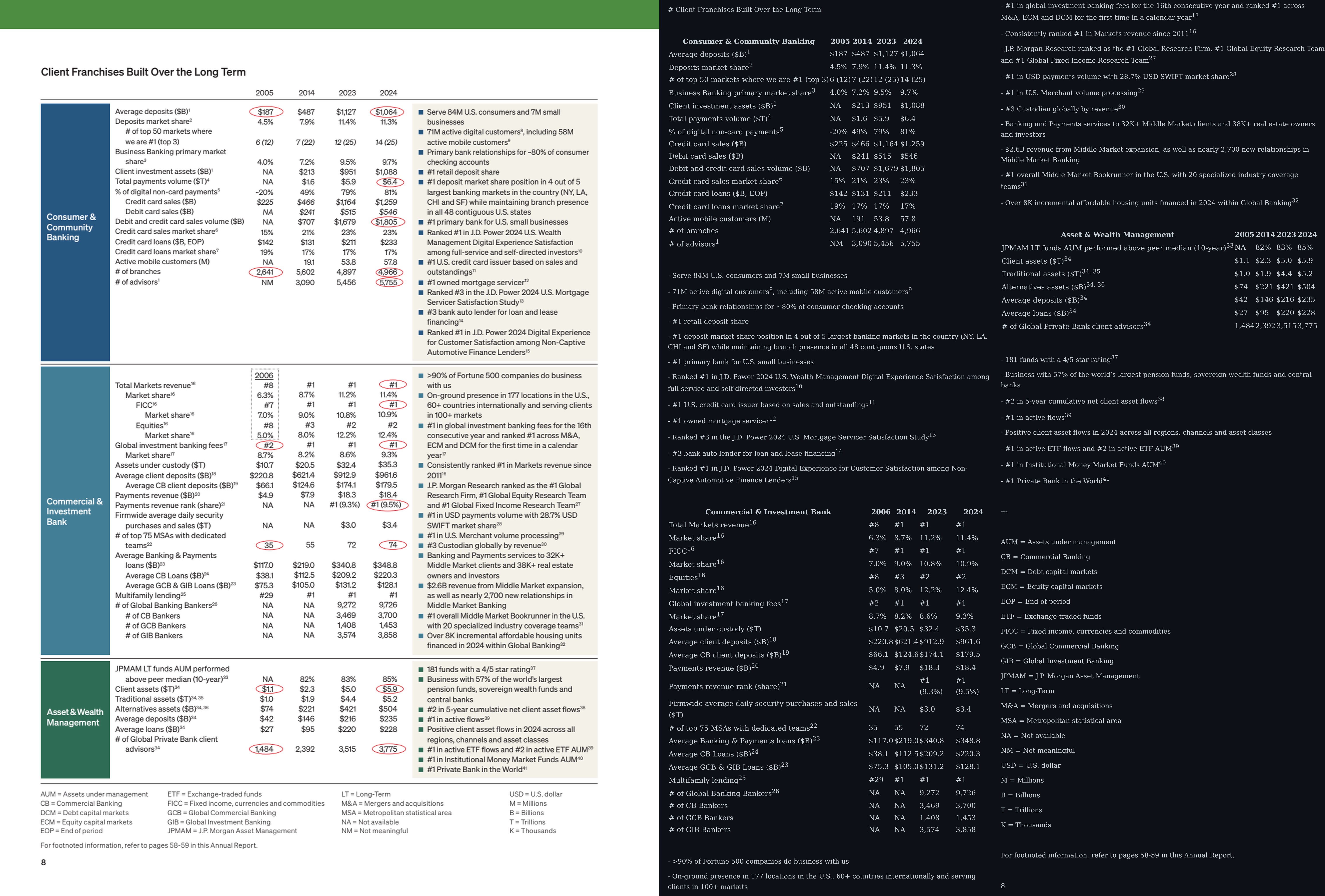}
    \caption{Complex table. Left: Original image. Right: Rendered transcription. Generated with LightOnOCR-2-1B.}
    \label{fig:placeholder}
\end{figure}
\begin{figure}[h!]
    \centering
    \includegraphics[width=1\linewidth]{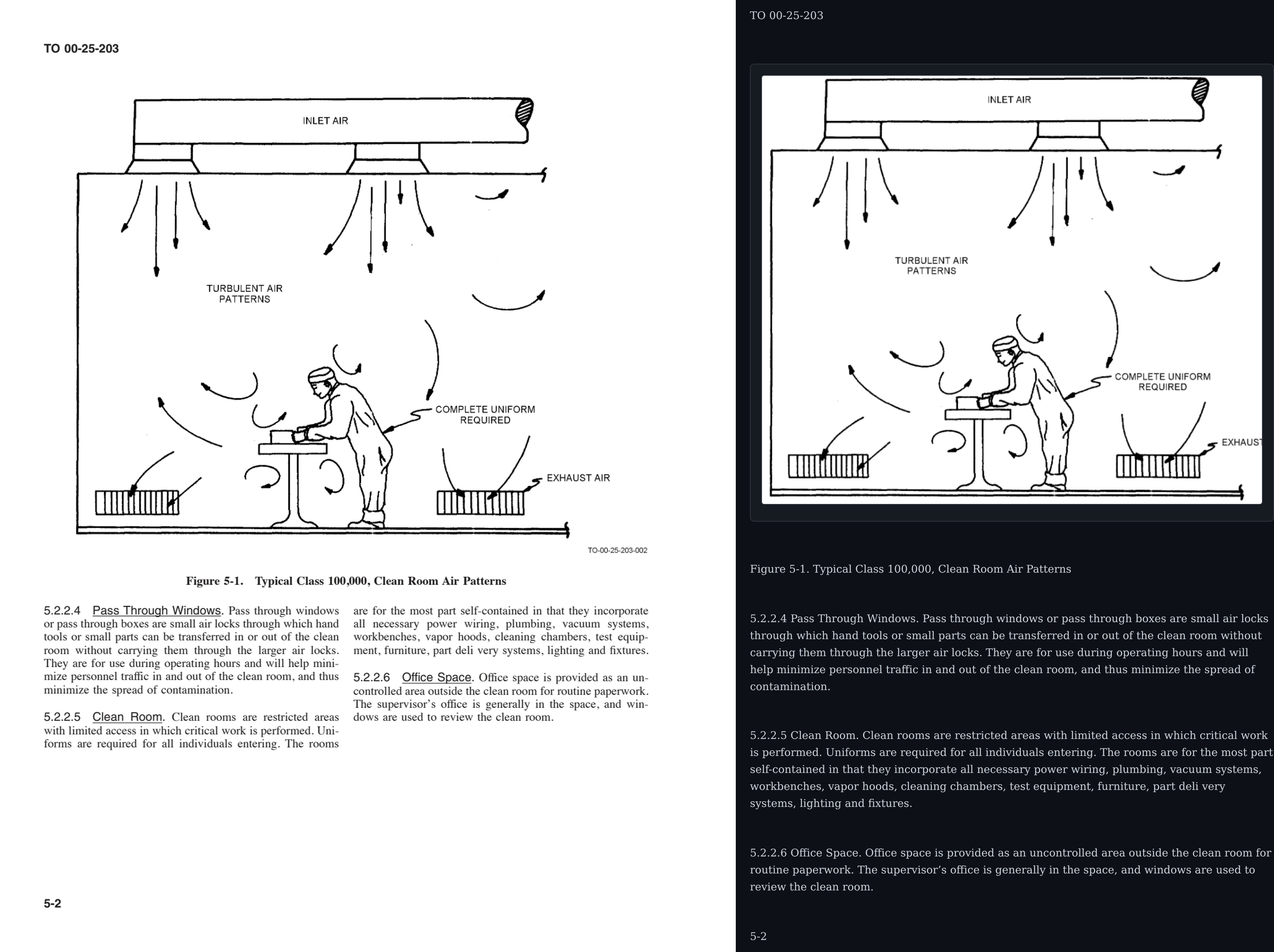}
    \caption{Example of bounding box generating models. Left: Original image. Right: Rendered transcription, with image crop corresponding to generated bounding box. Generated with LightOnOCR-2-1B-bbox. }
    \label{fig:placeholder}
\end{figure}
\begin{figure}[h!]
    \centering
    \includegraphics[width=1\linewidth]{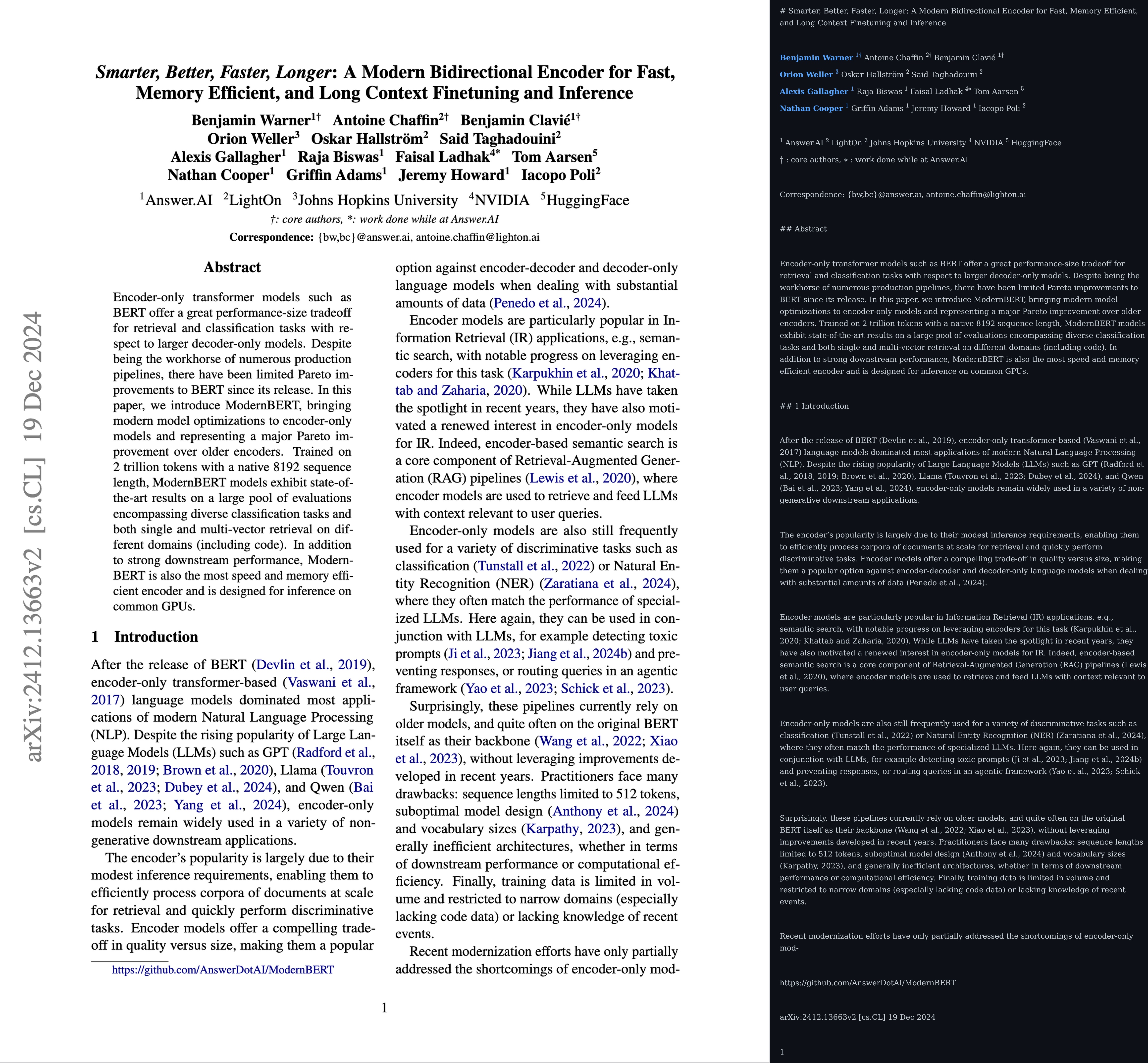}
    \caption{LightOnOCR-2-1B models excel at transcription of scientific documents. Left: Original image. Right: Rendered transcription. Generated with LightOnOCR-2-1B.}
    \label{fig:placeholder}
\end{figure}
\begin{figure}[h!]
    \centering
    \includegraphics[width=1\linewidth]{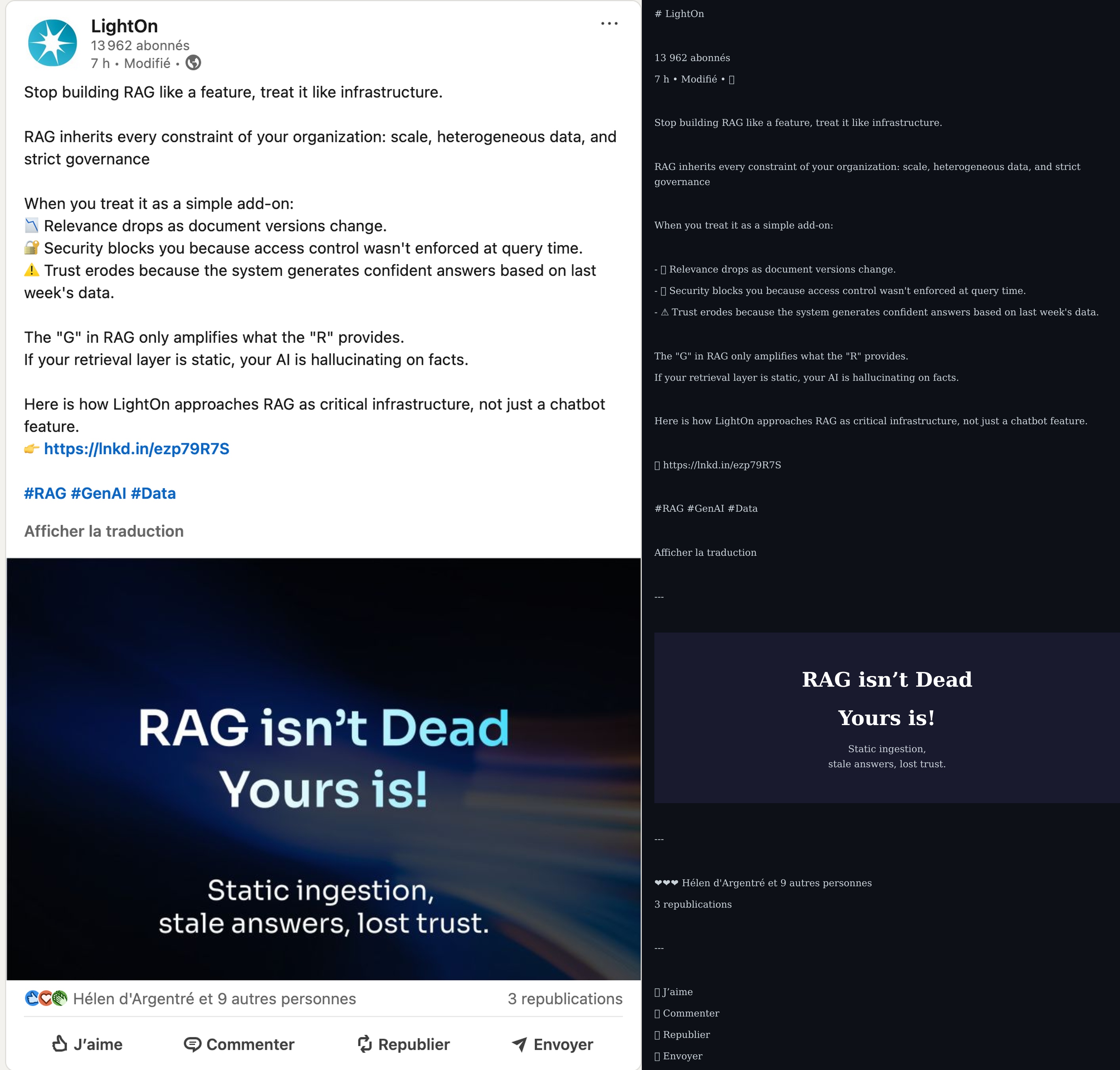}
    \caption{Example of transcription outside of distribution. Left: Original image. Right: Rendered transcription. Generated with LightOnOCR-2-1B.}
    \label{fig:placeholder}
\end{figure}
\begin{figure}[h!]
    \centering
    \includegraphics[width=1\linewidth]{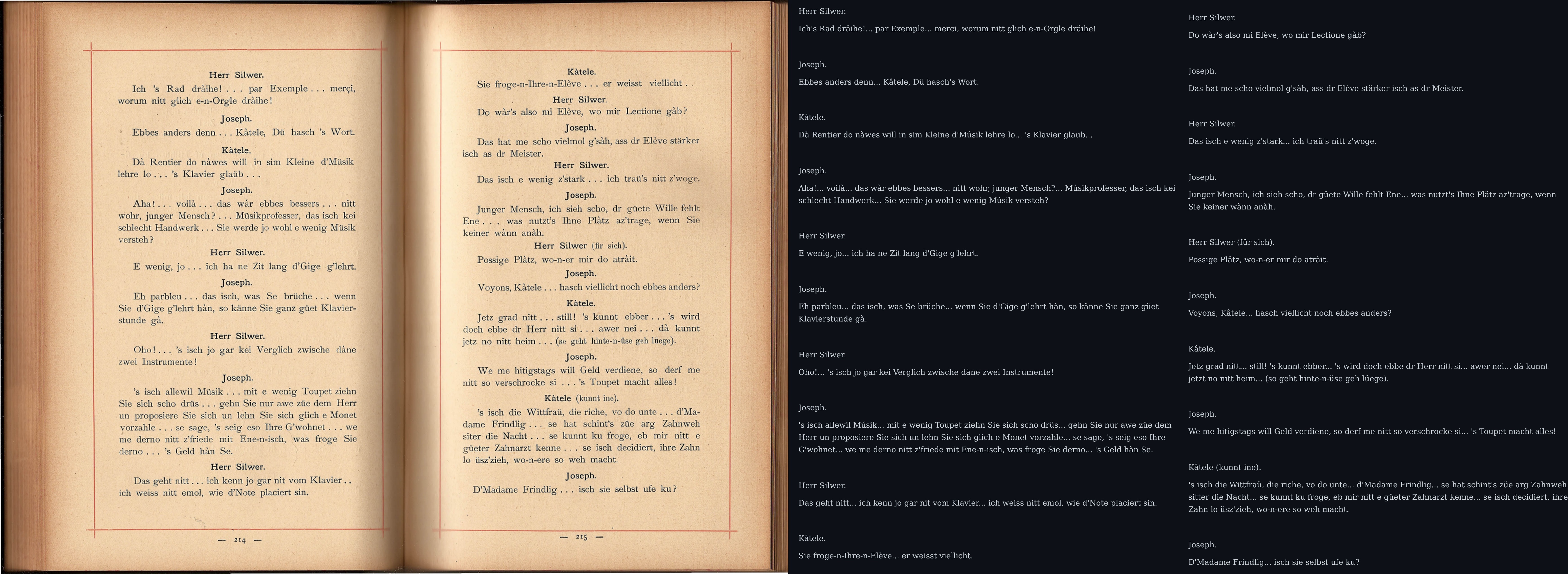}
    \caption{LightOnOCR-2-1B models were trained with a large portion of scanned documents which translates to improved performance on those types of files compared to v1.
    Left: Original image. Right: Rendered transcription. Generated with LightOnOCR-2-1B.
    Source: August Lustig, Sämtliche Werke – Zweiter Band (1909), p. 214–215. Wikimedia Commons, “ALustig\_SämtlicheWerke\_ZweiterBand\_page214\_215.pdf”, Public Domain (PDM 1.0)}
    \label{fig:placeholder}
\end{figure}
\clearpage

\end{document}